\documentclass[conference]{IEEEtran}
\usepackage{multirow}

\usepackage{cite}
\usepackage{amsmath,amssymb,amsfonts}
\usepackage{algorithmic}
\usepackage{graphicx}
\usepackage{textcomp}
\usepackage[table]{xcolor}
\usepackage{xcolor}
\usepackage{pgfplotstable}

\def\BibTeX{{\rm B\kern-.05em{\sc i\kern-.025em b}\kern-.08em
		T\kern-.1667em\lower.7ex\hbox{E}\kern-.125emX}}

\def\ccag#1{%
	\pgfmathsetmacro\calc{25}%
	\edef\clrmacro{\noexpand\cellcolor{green!\calc}}%
	\clrmacro%
	\ifdim \calc pt>50pt\color{black}\fi{#1}%
}

\def\ccar#1{%
	\pgfmathsetmacro\calc{25}%
	\edef\clrmacro{\noexpand\cellcolor{red!\calc}}%
	\clrmacro%
	\ifdim \calc pt>50pt\color{black}\fi{#1}%
}
\begin{document}

	\title{Assessing Risks of Biases in Cognitive Decision Support Systems
	}
	\author{\IEEEauthorblockN{ 
			{Kenneth Lai$^{1}$,} Helder C. R. Oliveira$^{1}$, Ming Hou$^{2}$, Svetlana N. Yanushkevich$^{1}$, and Vlad Shmerko$^{1}$ 
		}
		\IEEEauthorblockA{
			\textit{$^1$Biometric Technologies Laboratory, Department of Electrical and Computer Engineering,} \textit{ University of Calgary, Canada,} \\
			Web: http://www.ucalgary.ca/btlab, E-mail: \{kelai,helder.rodriguesdeol,syanshk,vshmerko\}@ucalgary.ca}\newline
		\textit{$^2$Defence Research and Development Canada (DRDC), Canada.} E-mail: ming.hou@drdc-rddc.gc.ca
	}
	
	\maketitle

	\thispagestyle{empty}

	\begin{abstract}
		
		Recognizing, assessing, countering, and mitigating the biases of different nature from heterogeneous sources is a critical problem in designing a cognitive Decision Support System (DSS). An example of such a system is a cognitive biometric-enabled security checkpoint. Biased algorithms affect the decision-making process in an unpredictable way, e.g. face recognition for different demographic groups may severely impact the risk assessment at a checkpoint. This paper addresses a challenging research question on how to manage an \emph{ensemble of biases}? We provide performance projections of the DSS operational landscape in terms of biases. A probabilistic reasoning technique is used for assessment of the \emph{risk of such biases}. We also provide a motivational experiment using face biometric component of the checkpoint system which highlights the discovery of an ensemble of biases and the techniques to assess their risks. 
	\end{abstract}
	
	

	\section{Introduction}\label{sec:}
	Cognitive Decision Support System (DSS) is an integrated part of various complex systems that are designed based on the automated cognition concept, which is based on the perception-action cycle. Examples of DSS include 
	unmanned aircraft systems \cite{hou2007intelligent, hou2014intelligent}, biometric-enabled security checkpoints \cite{yanushkevich2019cognitive}, supply chain management \cite{ojha2018bayesian}, and automated interviewing \cite{yanushkevich2019cognitive}. 
	Performance of cognitive DSS is evaluated in various dimensions:
	\begin{itemize}
		\item [$-$]Technical, e.g. false accept and false reject rate 
		\cite{lai2017bridging},
		\item [$-$]Social, e.g. public acceptance \cite{andreou2017identity},
		\item [$-$]Psychological, e.g. efficiency of human-machine interactions \cite{hu2018computational,hugenberg2013towards,montibeller2015cognitive}, and
		\item [$-$]Privacy and security domain, e.g. vulnerability of personal data 
		\cite{andreou2017identity,merler2019diversity,yanushkevich2019cognitive}. 
	\end{itemize}
	
\begin{figure}[!ht]
		\begin{center}
			\includegraphics[scale=0.35]{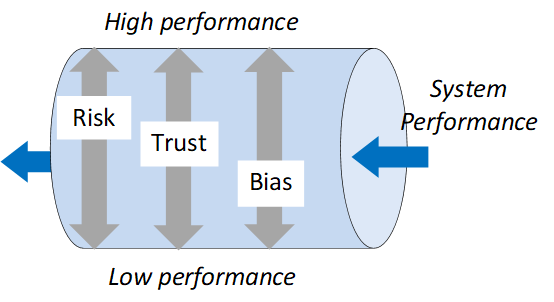}
			\vspace{-4mm}
		\end{center}
		\caption{Illustration of the problem:
			risk, trust, and bias are system regulators of technical, social, psychological, and security performance at various decision-making levels throughout various stages of human-machine and machine-machine interactions of cognitive DSS. 
		}\label{fig:Risk-Trust_Bias_Regulators}
	\end{figure}

	Taxonomy and regulators such as \emph{risk, trust}, and \emph{bias} are useful in the performance evaluation of complex dynamical systems (Fig. \ref{fig:Risk-Trust_Bias_Regulators}). For example, trust to the artificial intelligence (AI) interview assistant addresses the so-called AI (as well as cognitive) biases \cite{yanushkevich2019cognitive}; 
	risk and trust to DSS is related to various kinds of biases, e.g. in human identification based on face biometrics \cite{das2018mitigating,grother2019face} and social profiles \cite{andreou2017identity}. This paper focuses on risk and related assessment of an \emph{ensemble of biases} observed in a cognitive DSS, in particular, in a cognitive security checkpoint. Specifically, we are interested in the impact of biases on the DSS's performance.
	
	Many kinds of biases in biometric-based DSS were identified and studied. However, \textbf{dealing with an ensemble of biases is an open problem}. This is the focus of our interest. In our study, we model the DSS as a complex multi-state dynamic system \cite{yanushkevich2019cognitive,yanushkevich2019cognitive2}. Each state is characterized by a particular kind of bias. \textbf{The research question is how to fuse these biases in order to obtain the combined risk of biases and related trust estimators.} 
	
	Some preliminary results have been reported in the related areas.	
	In \cite{yao2013multi}, a trust bias interpreted as a prior knowledge has been incorporated into a non-probabilistic trust inference procedure. Three kinds of biases were distinguished: 1) a global bias, 2) a trustor bias, and 3) a trustee bias. A probabilistic network that is suited for approximate probabilistic confidence reasoning for trust inference problems has been developed in \cite{kuter2007sunny}. The DSS such as semi-automated cognitive checkpoint deals with \textbf{multiple (ensemble) of biases of different nature} from the following sources:
	
	\begin{footnotesize}
		\begin{center}
			\begin{parbox}[h]{0.95\linewidth} {
					\vspace{-2mm}
					\begin{center}
						\begin{eqnarray*}
							\textbf{\texttt{Bias}}
							\equiv \left\{
							\begin{tabular}{ll}
								\texttt{\small Human Perception}    & \cite{hugenberg2013towards,montibeller2015cognitive} ; \\
								\texttt{\small Artificial Intelligence} &\cite{yanushkevich2019cognitive} ;\\
								\texttt{\small Biometrics}       &\cite{das2018mitigating,grother2019face,merler2019diversity}; \\
								\texttt{\small Identity Management}   & \cite{andreou2017identity,yanushkevich2019cognitive}
							\end{tabular}    
							\right.
						\end{eqnarray*}
				\end{center}} 
			\end{parbox}
		\end{center}
	\end{footnotesize}

	A phenomenon of \emph{own-race bias} is well-known in psychology, the tendency to have better recognition for faces of one's racial ingroup rather than for racial outgroup faces \cite{hugenberg2013towards}, was recently confirmed in face recognition experiments \cite{das2018mitigating,grother2019face,merler2019diversity}. The source of additional cognitive biases is human-human interactions. The AI biases are introduced by intelligent support of human-machine interactions \cite{yanushkevich2019cognitive}. Finally, identity management bias was analyzed in multiple social profiles \cite{andreou2017identity}.
	
	While all these biases are different, they are probabilistic in nature because the evidence and information gathered to make a decision is always incomplete, often inconclusive, frequently ambiguous, commonly dissonant, conflicting, and
	has various degrees of believability. Trust and risk contribute to subject acceptance and rejection, respectively. Trust deviation impacts bias and vice versa. For example, by declining trust, one can increase bias. Trust is characterized by risks taken by a trustor in the presence of uncertainty.
	
	\textbf{The goal of this paper} is to develop a bias-specific projection of the DSS such as cognitive security checkpoints. The performance of the security checkpoint is a good indicator of technological progress. The checkpoint is a complex system that includes various sensors, biometrics and data processing, pattern recognition and decision-making components. It is also an indicator of public acceptance of the privacy transformation. \textbf{The contributions} of our study include: 1) a systematic view on the risk-trust-bias impact on the cognitive DSS performance, and 2) an approach to the ensemble of bias assessment and operation.

	This paper is organized as follows. After providing a background (Section \ref{sec:Basics}), {our approach is introduced in Section \ref{sec:Taxonomical-projection}. The details are highlighted in Section \ref{sec:experiment} via an experiment.} Section \ref{sec:Summary} concludes the paper.

	\section{Background}\label{sec:Basics}
	
	Bias is an application-specific phenomenon in the real-world: 
	\begin{itemize}
		\item [$-$]In statistics, bias refers to the tendency of a measurement process to systematically over- or under-estimate the value of a population parameter. 
		\item [$-$]In psychology, bias is a well-defined phenomenon: 
		e.g., a customer bias is a deviation of the forecast from the true expectation. 
		\item [$-$]In biometric systems, biases occur, for example, in the form of a systematic difference between facial recognition of individuals belonging to different demographic groups 
		(performance bias), a systematic decision-making mistake (prediction bias), and a systematic mis-classification (classification bias).
		\item [$-$]In Artificial Intelligence (AI), any machine learning-based assessment  has a bias. 
		The key question is how to teach an AI assistant to act within the ethical and legal guidelines. Fairness detection can be referred to as AI bias. 
	\end{itemize}
	
	Identifying and mitigating bias is essential to build {trust} between humans and machines. The machines can learn and assess the related {risks}. Risk of the biased AI judgment is a focus of interest in all AI applications, e.g. in medicine \cite{gates2018technology}.
	
	Various characteristics are needed for the performance evaluation of the DSS. For example, the commonly used false reject rate and false accept rate in decision-making are affected by the level of the individual's satisfactory and social acceptance of profiling technologies. Other research challenges include identifying biases in human biometric and behavior recognition, in confidence estimation, AI trustworthiness, and belief inference in machine reasoning.
	Each of these extensions is considered over specific operational landscapes. For example, bias cannot be propagated through system states, but risk and trust of bias can be represented by probability distributions and quantitatively assessed as probabilities. 
	
	{Risk} is a function of (a) an adverse impact, or magnitude of the harm, that would arise if the circumstance or event occurs; and (b) a likelihood of occurrence, \texttt{Risk$=F$(Impact, Probability)} \cite{nist2017security}. 
	
	
	\section{Bias taxonomy and operational landscape}\label{sec:Taxonomical-projection}
	
	There are three phases of bias analysis in the DSS: 
	1) Bias identification, e.g, what kind of biases are manifested in a system?
	2) Bias assessment, e.g. a unified metric for different kinds of biases, such as the risk of bias; and
	3) Bias operation, e.g. fusion of risks of biases.
	Recent studies are focused on bias identification such as demographic bias in facial recognition \cite{das2018mitigating,grother2019face,merler2019diversity} and AI bias \cite{gates2018technology}. This is a starting point in cognitive DSS development. Given the DSS multi-state model, the key requirements to the identified bias are assessment metric that provides bias operations such as propagation, adjustment, fusion, and prediction. 
	
\begin{figure}[!ht]
		\begin{center}
			\includegraphics[width=0.5\textwidth]{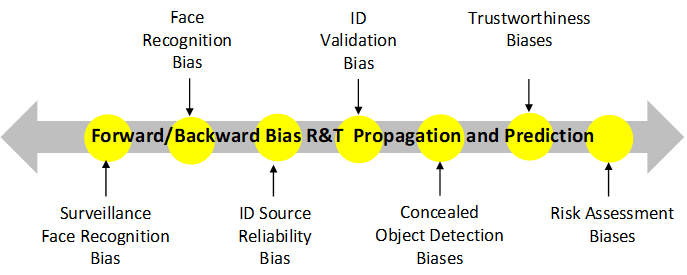}
			\caption{Taxonomical projection of the cognitive security checkpoint in terms of biases: biases are propagated between states, and at each state, their status is analyzed, adjusted, fused, and predicted.}
			\label{fig:Taxonomical_projection}
		\end{center}
	\end{figure}

	
	The aforementioned aspects of bias analysis are introduced in Fig. \ref{fig:Taxonomical_projection} for the particular case of the DSS, cognitive security checkpoint. The traveler's identity management process is implemented as a process that goes through multiple states \cite{yanushkevich2019cognitive,yanushkevich2019cognitive2}.  Each state is characterized by a specific bias such as bias in face recognition when using surveillance and ID check, the bias of ID source reliability, etc. Statistics of these biases are being used for machine learning. The biases are mostly represented by the tailed probability distributions. A unified metric of bias that we consider in this study is the risk of bias and the related trust in the technology that is biased.

	Given the risk of a particular bias, various operations can be available with this risk value. 
	For example, the  \emph{forward propagation} reflects a process that relates the causes to their respective effects. The bias \emph{backward propagation} reflects the bias assessment that traces the effects to their respective causes. Forward and backward bias propagation provide a systematic evaluation of the biases in traveler profiling or risk assessment. These biases are caused by various threats, hazards, and concerns, and are useful in deriving the cost-effective measures for lowering these risks to an acceptable level. We suggest that various techniques from related areas can be adopted for this purpose, in particular, causal relationships of risk factors and their propagation \cite{feng2014security}, risk propagation in supply network \cite{garvey2015analytical,ojha2018bayesian}, as well as trust prediction and belief propagation \cite{zhang2014trust}.

	\section{Motivational experiment}\label{sec:experiment}
	
	Fig. \ref{fig:Biases} illustrates various operations on biases. Bias is represented in metrics that must be compatible with metrics of risk and trust. The latter is represented by probability distributions of measures of the AI decision performance, decision reliability, confidence, credibility, and trustability. These values, in terms of posterior probabilities, can be inferred based on the prior probabilities and the current observation or conditions. The mechanism we apply for such inference is Bayesian, also called belief networks.

	The motivations of an experiment are as follows:
	1) Explain the details of the ensemble bias assessment, and
	2) Introduce the essentials of reasoning mechanism.
	Our motivational experiment addresses multiple biases in a cognitive security checkpoint (Fig. \ref{fig:Taxonomical_projection}) such as ``ID Reliability Bias'', ``ID Validation Bias'', and ``Trustworthiness Bias''.
	Among various candidates of biases, we will consider ``Face Recognition Bias''.
	Few results on demographic bias in facial recognition have been recently reported, in particular, in \cite{das2018mitigating,grother2019face,merler2019diversity}.
	Our motivational experiment aims at highlighting the practical details of assessing an ensemble of biases. 
	
	\subsection{Causal model}
\begin{figure}[!ht]
		\begin{center}
			\includegraphics[width=0.35\textwidth,interpolate]{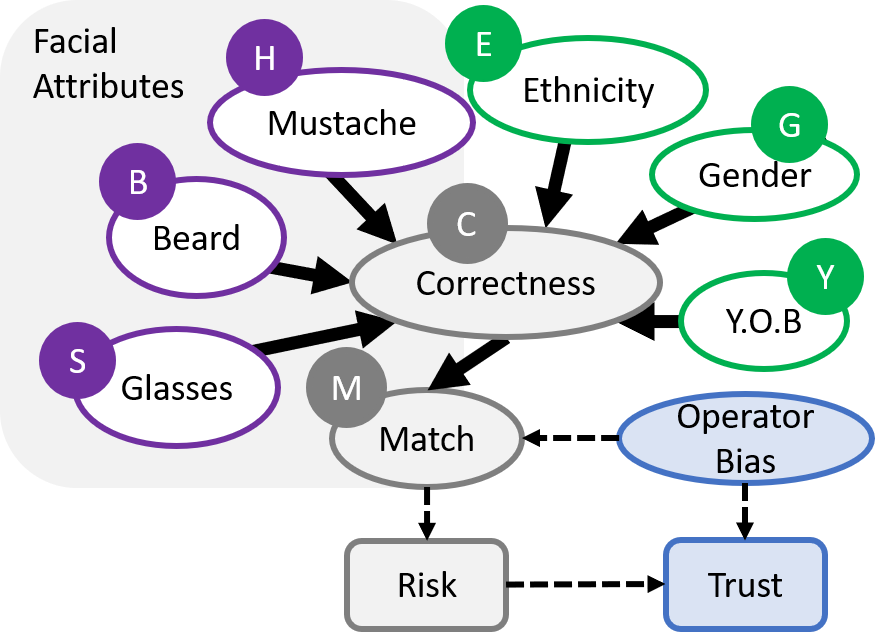}
			\caption{ A simplified causal network of biases for facial recognition.  Risk is derived based on the results of the ``Match'', and Trust is affected by the Operator's Bias }
			\label{fig:Biases}
		\end{center}
	\end{figure}

	The causal network shown in Fig. \ref{fig:Biases} describes how the quality of facial recognition can be compromised by various facial attributes that are ``biased'' based on the year-of-birth (YOB) $Y$, gender $G$, ethnicity $E$, mustache $H$, beard $B$, and glasses $S$. The parent nodes to the ``Correctness'' node represents the bias attributed to face recognition. The ``Correctness'' node presents the probability of the neural network in predicting a positive (genuine subject) or negative (imposter) identity, whereas the ``Match'' node determines whether the positive or negative prediction matches the ground truth label.
	
	\subsection{Formalization}
	
	The risk of the decision is evaluated based on the results of the ``Match'' node. The Risk  affects the perceived Trust, which is also affected by the Operator's Bias (such as trained operator, inexperienced operator, perceived AI trustworthiness, and other human factors). The Operator then can use their judgment to adjust the parameters of the matcher in order to improve the decision outcome, thus implementing the ``Perception-Action'' cycle of the cognitive DSS. The adjustment action, in turn, will affect the Operator's trust in the DSS decision. It should be noted that the trust assessment would require the quantitative representation of the human factor (Operator) bias, which was not evaluated in this experiment. Therefore, we will focus on the risk assessment in the experiment described below. 
	
	Risk is estimated as \texttt{Risk$=F$(Impact, Probability)} which relates to the error rates of the system, specifically the false non-match rate (FNMR) and false match rate (FMR). At a high level of abstraction (e.g. ignoring metric and dependencies), risk of a particular bias $\texttt{Risk}_\text{\slshape Bias}$ is defined as follows:
	
	\vspace{-4mm}
	\begin{small}
		\begin{eqnarray} \label{eq:riskb}
		\texttt{Risk}_\text{\slshape Bias}&=& \underbrace{\texttt{Impact}_{\text{\slshape FMR}}}_\text{Cost of a FMR} \times \underbrace{\texttt{Error}_{\text{\slshape FMR}}}_{\text{FMR}} \nonumber\\
		&+&\underbrace{\texttt{Impact}_{\text{\slshape FNMR}}}_\text{Cost of a FNMR} \times \underbrace{\texttt{Error}_{\text{\slshape FNMR}}}_{\text{FNMR}}
		\end{eqnarray}
	\end{small}
	
	\vspace{-3mm}
	For example, given the scenario of a security checkpoint, the FMR is related to a wrongly granted access, while the FNMR contributes to travelers' inconvenience. The impact of the FMR is a breach of security which, given this scenario, should have a high impact. The impact of the FNMR is a negative user experience which is rather of a low impact. Using this scenario, we assign a 10:1 ratio, that is $\texttt{Impact}_{FMR}=10$ and $\texttt{Impact}_{FNMR}=1$. Given the Year-of-Birth (YOB) attribute, the risk for individuals born in the 1930s is computed as follows: $10 \times 0.0208 + 1 \times 0.0012 = \fbox{0.2092}$.
	
	The ensemble risk bias in identifying (matching) a particular individual $\texttt{Risk}_{\text{\slshape Bias}}\texttt{(Ensemble)}$ is assessed as the sum of the risk biases according to his/her attributes:
	
	\vspace{-3mm}
	\begin{small}
		\begin{eqnarray} \label{eq:risk}
		\texttt{Risk}_{\text{\slshape Bias}}\text{(\slshape Ensemble)}&=& \sum_{\text{\slshape Bias}=1}^{N} \texttt{Risk}_\text{\slshape Bias}
		\end{eqnarray}
	\end{small}
	where $\text{\slshape Bias}$ represents one of the attributes, $Y,G,E,H,B,$ and $S$ in Fig. \ref{fig:Biases}, $N=6$.

	\subsection{Experimental setup}
	
	For this experiment, we demonstrate biases using the FERET face database. The database was collected between August 1993 and July 1996 and contains a total of 14,126 images, including 1199 individual subjects \cite{phillips1998feret}. The FERET database was chosen because of its detailed meta-data information which includes information such as gender, year-of-birth, ethnicity, and pose.
	
	The typical performance of face recognition includes accuracy, FNMR, and FMR and are defined as $\mathrm{Accuracy} = \frac{\mathrm{TP}+\mathrm{TN}}{\mathrm{FP}+\mathrm{FN}+\mathrm{TP}+\mathrm{TN}}$, $\mathrm{FNMR} = \frac{\mathrm{FN}}{\mathrm{FN}+\mathrm{TP}}$, and $\mathrm{FMR} = \frac{\mathrm{FP}}{\mathrm{FP}+\mathrm{TN}}$, where TP are true positives (correct prediction identity), TN are true negatives (correct prediction of imposter), FP are false positives (incorrect prediction of identity), FN are false negatives (incorrect prediction of imposter).
	
	The features used for face identification is extracting using a pre-trained Resnet50 convolutional neural network. The features are then passed through a fully-connected layer and softmax activation to perform identification. The model is trained using stochastic gradient descent with a learning rate $lr=0.0001$ and a momentum of 0.9 for 100 epochs.
	
	\subsection{Results}
	
	The overall identification accuracy, FNMR, and FMR for the FERET dataset are reported in Table \ref{tab:recognition}. The results are separated based on the biases such as gender, year-of-birth, ethnicity, and facial attributes (glasses, beard, and mustache). The performance measures are evaluated based on the Rank-1 results. Given a baseline accuracy of $91.63\%$, \textbf{positive biases are highlighted in green while negative biases are highlighted in red}. For example, observing the attribute year-of-birth, a significant bias is shown as the year-of-birth increases, specifically significant bias against younger individuals. \textbf{An accuracy difference of $\mathbf{17.65\%}$ between those born in the 1920s and the 1980s}.  The FNMR and FMR for each bias are computed based on the bias nodes of the causal network shown in Fig. \ref{fig:Biases}.
	
	\begin{table}[!htb]
		\centering
		\caption{Performance of the face identification process in terms of the accuracy, FNMR, and FMR (FERET dataset).}\label{tab:recognition}
		\begin{tabular}{@{}c|ccc@{}}
			Method	&	Accuracy	& FNMR	&	FMR	\\
			\hline
			Baseline	&	0.9163	&	0.0941	&	0.0001	\\
			\hline							
			Female	&\ccar{0.8908	}&	0.1233	&	0.0003	\\
			Male	&\ccag{0.9308	}&	0.0749	&	0.0001	\\
			\hline							
			1920s	&\ccag{1.0000	}&	0.0000	&	0.0000	\\
			1930s	&	0.9697	&	0.0208	&	0.0012	\\
			1940s	&	0.9667	&	0.0443	&	0.0004	\\
			1950s	&	0.9513	&	0.0625	&	0.0003	\\
			1960s	&	0.9122	&	0.1020	&	0.0004	\\
			1970s	&	0.8873	&	0.1124	&	0.0003	\\
			1980s	&\ccar{	0.8235	}&	0.1875	&	0.0105	\\
			\hline							
			Asian	&	0.9424	&	0.0664	&	0.0003	\\
			Black-or-African-American	&	0.9195	&	0.0984	&	0.0014	\\
			Hispanic	&\ccar{0.8254	}&	0.1981	&	0.0033	\\
			Native-American	&\ccag{1.0000	}&	0.0000	&	0.0000	\\
			Other	&\ccag{1.0000	}&	0.0000	&	0.0000	\\
			Pacific-Islander	&\ccag{1.0000	}&	0.0000	&	0.0000	\\
			White	&	0.9119	&	0.0965	&	0.0002	\\
			\hline							
			No Glasses	&\ccag{0.9214	}&	0.0923	&	0.0001	\\
			Glasses	&\ccar{0.8859	}&	0.1028	&	0.0009	\\
			\hline							
			No Beard	&\ccar{0.9134	}&	0.0976	&	0.0001	\\
			Beard	&\ccag{0.9615	}&	0.0357	&	0.0007	\\
			\hline							
			No Mustache	&\ccar{0.9144	}&	0.0947	&	0.0001	\\
			Mustache	&\ccag{0.9328	}&	0.0891	&	0.0008	\\
			
		\end{tabular}
	\end{table}

	Table \ref{tab:subjects} illustrates the effect of biases on the accuracy of the chosen classifier in identifying 11 randomly selected subjects with their associated attributes. The top 3 identity predictions (with their respective score values) are shown in the last three columns. \textbf{The first 8 subjects are samples of correctly identified subjects, while the last 3 subjects are incorrectly identified subjects when using only the top/rank-1 prediction.} The highlighted results indicate a match to the ground truth label. The score values indicate that if we choose to use a threshold-based method of classification, it is not possible to decisively separate the genuine and imposter scores. 
	Only the top 5 scores of each prediction are used for creating the genuine and imposter scores. Scores, representing the similarity between faces, are generated from the penultimate layer of the neural network. Similar/identical faces generate large scores while unrelated faces produce low/negative scores. Through softmax normalization, the sets of scores are converted into probabilities. Due to the exponential nature of softmax normalization, lower scores, when converted to probabilities, are effectively reduced to zero, and thus, should not influence the decision-making process.
	
	\begin{table*}[!htb]
		\centering
		\caption{Bias phenomenon of the facial recognition process: attributes for random Subjects (FERET dataset).}\label{tab:subjects}
		\begin{scriptsize}
			\begin{tabular}{r|ccc|ccc|rrr}
				\multirow{2}{*}{Subject}	&	Year of	&	\multirow{2}{*}{Gender}	&	\multirow{2}{*}{Ethnicity}	&	\multicolumn{3}{c|}{Facial Features}	& \multicolumn{3}{c}{Top Subject Predictions}	\\
				
				&	Birth	&		&	&	Glasses	&	Beard	&	Mustache	&	First	 (	Score	)	&	Second	 ( Score		)	&	Third	 (	Score	)		\\
				\hline																															
				15	&	 1970s	&	 Male	&	 Asian	&	 False	&	 False	&	 False	& \ccag{	15	 }(	11.8175	)	&	913	 (	7.9727	)	&	612	 (	7.5015	)	\\
				189	&	 1970s	&	 Female	&	 Asian	&	 False	&	 False	&	 False	& \ccag{	189	 }(	15.7484	)	&	434	 (	8.5322	)	&	566	 (	6.9052	)	\\
				143	&	 1970s	&	 Female	&	 Hispanic	&	 False	&	 False	&	 False	& \ccag{	143	 }(	12.1884	)	&	402	 (	6.6154	)	&	763	 (	5.4235	)	\\
				295	&	 1980s	&	 Female	&	 Black-or-African-American	&	 False	&	 False	&	 False	& \ccag{	295	 }(	14.4057	)	&	364	 (	6.6878	)	&	808	 (	6.2970	)	\\
				561	&	 1940s	&	 Male	&	 Asian	&	 False	&	 False	&	 False	& \ccag{	561	 }(	10.4680	)	&	187	 (	6.6516	)	&	695	 (	6.0315	)	\\
				917	&	 1940s	&	 Female	&	 Hispanic	&	 False	&	 False	&	 False	& \ccag{	917	 }(	10.5805	)	&	251	 (	6.8186	)	&	972	 (	6.5261	)	\\
				948	&	 1950s	&	 Male	&	 White	&	 False	&	 False	&	 True	& \ccag{	948	 }(	10.7142	)	&	330	 (	5.4853	)	&	962	 (	5.1746	)	\\
				684	&	 1930s	&	 Male	&	 Asian	&	 False	&	 False	&	 False	& \ccag{	684	 }(	7.7132	)	&	501	 (	6.3049	)	&	775	 (	5.7674	)	\\
				\hline
				0	&	 1940s	&	 Male	&	 White	&	 True	&	 False	&	 False	&	457	 (	8.0854	)	& \ccag{	0	 }(	6.7735	)	&	348	 (	6.3287	)	\\
				774	&	 1950s	&	 Female	&	 White	&	 False	&	 False	&	 False	&	704	 (	9.4579	)	&	801	 (	7.3339	)	&	942	 (	7.0282	)	\\
				255	&	 1960s	&	 Female	&	 White	&	 False	&	 False	&	 False	&	924	 (	8.8947	)	& \ccag{	255	 }(	7.0531	)	&	270	 (	6.9886	)	\\
			\end{tabular}
		\end{scriptsize}
	\end{table*}


	\subsection{Reasoning on risk of bias}
	
	The model for reasoning upon the probability distribution of matching scores and the chosen biases is a probabilistic graphical model called Bayesian, or belief, network is shown in Fig. \ref{fig:Biases}. This is a causal network with the assigned Conditional Probability Tables (CPTs). Bayesian inference is applied to the CPTs to update the probability given evidence of various biases. 
	Example of calculation the prior probabilities for the parent nodes in the belief network (Fig. \ref{fig:Biases}) is as follows:
	
	\begin{footnotesize}
		\begin{center}
			\begin{tabular}{cc}
				\begin{tabular}{c|cc}
					& \multicolumn{2}{c}{Gender} \\
					$G$ & Male & Female \\
					\hline
					\(\Pr(G)\) & 0.6383 & 0.3617 \\
				\end{tabular} & 
				\begin{tabular}{c|cc}
					& \multicolumn{2}{c}{Wear Glasses} \\
					$S$ & True & False\\
					\hline
					\(\Pr(S)\) & 0.1425 & 0.8575	
				\end{tabular} 
			\end{tabular} 
		\end{center}	
	\end{footnotesize}

	These prior probabilities are vital for calculating the posterior probabilities representing the impact of biases on the resulting matching rate. It should be noted that for better dealing with uncertainty and conflicts, more advanced metrics rather than point probabilities as in Bayesian networks can be chosen to be applied in the causal network. These metrics of uncertainty include interval probabilities, fuzzy probabilities, as well as Demster-Shafer and Dezert-Smaradache extensions \cite{yanushkevich2019cognitive2}.
	
	Any bias scenario can be represented by a causal network, e.g. Bayesian network. 
	Using the reported performance from Table \ref{tab:recognition} for risk estimation and an impact factor of 1 for both FNMR and FMR, the baseline risk of bias is computed according to Equation (\ref{eq:risk}) as 
	$1\cdot 0.0941+1\cdot 0.0001=\fbox{0.0942}$.
	Let the attribute such as gender be known. Then the risk of bias is evaluated as follows:
	\vspace{-5mm}
	
	\begin{small}
		\begin{eqnarray*}
			\texttt{Risk}_\text{\slshape Bias}(\text{\slshape Female})&=&
			\texttt{Impact}_{\text{\slshape FMR}} \times \texttt{Error}_{\text{\slshape FMR}}^{\text{\slshape Female}}\\
			&+&{\texttt{Impact}_{\text{\slshape FNMR}}} \times \texttt{Error}_{\text{\slshape FNMR}}^{\text{\slshape Female}}\\
			&=&1\times 0.1233+1\times 0.0003=\fbox{0.1236};\\
			\texttt{Risk}_\text{\slshape Bias}(\text{\slshape Male})&=&
			\texttt{Impact}_{\text{\slshape FMR}} \times \texttt{Error}_{\text{\slshape FMR}}^{\text{\slshape Male}}\\
			&+&{\texttt{Impact}_{\text{\slshape FNMR}}} \times \texttt{Error}_{\text{\slshape FNMR}}^{\text{\slshape Male}}\\
			&=&
			1\times 0.0749+1\times 0.0001=\fbox{0.0750}
		\end{eqnarray*}
	\end{small}
	
	\vspace{-5mm}
	The risk associated with females (0.1236) is higher than males (0.0750) due to the recognition model being gender-biased. This can be reasoned due to a larger representation of males in the dataset, as evident in the prior node for gender distribution.
	
	To evaluate the perceived trust of the Operators in the matcher decision, the Operator's bias shall be taken into account. As well, based on the assessed risk, the Operator can take action in adjusting the parameters of the matcher such as the decision threshold, since the Operator wishes to improve the trustworthiness of the decision. Another cycle of inference can be performed, while this time the risk is interpreted as evidence of the existing bias. This ``Perception-Action'' cycle of the cognitive DSS will lead to continuous re-assessment of the risk and trust in the DSS decision.

	\section{Summary and conclusion}\label{sec:Summary}
	
	Our study addresses the probabilistic reasoning as a mechanism for assessing an ensemble of biases aiming at their further propagation, adjustment, fusion, and prediction. In real-world cognitive DSS scenarios, each risk of bias in the ensemble is introduced by a probability distribution function (pdf) rather than CPTs as considered in the motivational example. The future research avenue here would be to perform a fusion of pdf(s) using the standard copula technique that shall result in the joint pdf. Marginalization will result in conditional pdf(s) that are assigned to the nodes of the causal network. The followed up reasoning mechanism will allow for inferences of risks and trust in the causal model.
	
	\section*{Acknowledgments}
	\begin{small}
		This Project was partially supported by 
		Natural Sciences and Engineering Research Council of Canada (NSERC) through grant ``Biometric-Enabled Identity Management and Risk Assessment for Smart Cities'', and  the Department of National Defence’s Innovation for Defence Excellence and Security (IDEaS) program, Canada. 
	\end{small}
	
		{\small
			\bibliographystyle{IEEEtran}
			\bibliography{bias}
		}

\end{document}